\documentclass[10pt,twocolumn,letterpaper]{article}

 \usepackage[pagenumbers]{cvpr} 

\usepackage[accsupp]{axessibility}  

\newcommand{\disclaimer}{
    \vspace{-1.5cm}
    \begin{center}
        \normalsize\textit{To appear in the Proceedings of the IEEE/CVF Conference on Computer Vision and Pattern Recognition (CVPR) Workshops, 2025}        
    \end{center}
    \vspace{1.0cm}
}

\definecolor{cvprblue}{rgb}{0.21,0.49,0.74}
\usepackage[pagebackref,breaklinks,colorlinks,allcolors=cvprblue]{hyperref}
\usepackage{svg}

\def\paperID{*****} 
\def\confName{CVPR}
\def\confYear{2025}

\title{\disclaimer Learned Lightweight Smartphone ISP with Unpaired Data}

\author{Andrei Arhire\\
Faculty of Computer Science\\
Alexandru Ioan Cuza University of Iasi\\
{\tt\small andrei.arhire@student.uaic.ro}
\and
Radu Timofte\\
Computer Vision Lab, CAIDAS \& IFI \\
University of W\"urzburg\\
{\tt\small radu.timofte@uni-wuerzburg.de}
}

\begin{document}
\maketitle
\begin{abstract}

The Image Signal Processor (ISP) is a fundamental component in modern smartphone cameras responsible for conversion of RAW sensor image data to RGB images with a strong focus on perceptual quality. Recent work highlights the potential of deep learning approaches and their ability to capture details with a quality increasingly close to that of professional cameras. A difficult and costly step when developing a learned ISP is the acquisition of pixel-wise aligned paired data that maps the raw captured by a smartphone camera sensor to high-quality reference images.

In this work, we address this challenge by proposing a novel training method for a learnable ISP that eliminates the need for direct correspondences between raw images and ground-truth data with matching content. Our unpaired approach employs a multi-term loss function guided by adversarial training with multiple discriminators processing feature maps from pre-trained networks to maintain content structure while learning color and texture characteristics from the target RGB dataset. Using lightweight neural network architectures suitable for mobile devices as backbones, we evaluated our method on the Zurich RAW to RGB and Fujifilm UltraISP datasets. Compared to paired training methods, our unpaired learning strategy shows strong potential and achieves high fidelity across multiple evaluation metrics. The code and pre-trained models are available at \url{https://github.com/AndreiiArhire/Learned-Lightweight-Smartphone-ISP-with-Unpaired-Data}.

\end{abstract}    
\section{Introduction}
\label{sec:intro}

The transformation steps required for the image signal to pass through, starting from the CMOS sensor readings, on its way to reaching the refined RGB image seen by the users, are entirely handled by the Image Signal Processor (ISP). Several of such processing stages include denoising, demosaicing, color consistency, gamma correction, and compression. In traditional ISPs, these steps are hand-crafted and applied sequentially. Consequently, they propagate a small error through the processing chain, leading to a degradation of the results.

Multiple tasks performed by ISP have been addressed individually through deep learning with great outcomes. In recent years, the idea of creating a deep neural network capable of outperforming conventional ISP has received growing attention~\cite{ignatov2020replacingmobilecameraisp,ignatov2022microispprocessing32mpphotos}. Considering the trade-off between latency and performance, developing a learned ISP intended to run on edge devices has been the subject of various challenges and currently represents an active field of research. 

A learnable ISP has the ability to partially overcome specific constraints, such as a small sensor and limited optical system, reducing the perceptual gap between smartphone cameras and professional DSLRs. To get the best results, the model is expected to be trained using paired pixel-wise aligned data. In practice, such data sets are difficult to obtain and must be collected individually for each new camera, since the characteristics of one camera directly impact the raw data. A recent solution to this challenge is Rawformer~\cite{perevozchikov2024rawformerunpairedrawtorawtranslation}, which proposes a state-of-the-art unsupervised method to translate the raw training data set from a specific camera domain to the target camera domain. However, a learned ISP with unpaired data could potentially provide a more accurate representation of the ground truth as it works directly with the original data.

Inspired by the WESPE work of Ignatov~\etal~\cite{ignatov2018wespeweaklysupervisedphoto}, we introduce an unpaired learning approach to train a learnable ISP. To ensure minimal latency on edge devices, our experiments primarily use the model architecture developed by the winner of the Mobile AI \& AIM 2022 Learned Smartphone ISP Challenge~\cite{ignatov2022learnedsmartphoneispmobile}. In our proposed pipeline, the model is trained using a multi-term loss function with dedicated components for content, color, and texture. To capture various characteristics of the statistical distribution of the target dataset, three discriminators are used during training. Guided by relativistic adversarial losses, the model learns to enhance color fidelity and perceptual realism, while structural consistency is preserved through a self-supervised loss. Our approach is evaluated on two real-world RAW-to-RGB datasets, Zurich RAW-to-RGB~\cite{ignatov2020replacingmobilecameraisp} and Fujifilm UltraISP~\cite{ignatov2022microispprocessing32mpphotos}. The generated images achieve fidelity scores (PSNR, SSIM~\cite{wang2004image}, MS-SSIM~\cite{wang2003multiscale}) comparable to those obtained through paired training, while maintaining a favorable perceptual quality (LPIPS score~\cite{zhang2018unreasonableeffectivenessdeepfeatures}).

The remainder of the paper is organized as follows. Section~\ref{sec:related_work} reviews the related work. Section~\ref{sec:proposed_method} introduces the proposed methods.
Section~\ref{sec:experiments} describes the experimental setup and the achieved results, while the conclusions are drawn in Section~\ref{sec:conclusion}.

\section{Related Work}
\label{sec:related_work}

A learnable ISP is trained to translate the image from the RAW format to an RGB domain with superior visual quality, refined for human perception. Typically, this is achieved by training on paired and pixel-wise aligned image patches using RAW data from a particular camera sensor and images from a high-quality DSLR camera.  

PyNET~\cite{ignatov2020replacingmobilecameraisp} is one of the first learnable ISPs that achieves the efficiency of the Huawei P20 commercial pipeline ISP and obtained superior results in the evaluation of mean opinion scores (MOS). It relies on an inverse pyramidal CNN architecture, which processes input on different scales. On each scale, specific features are learned with dedicated loss functions. In follow-up research, PyNET-CA~\cite{Kim_2020} improves performance by incorporating a channel-attention mechanism. Mobile-suitable variants, including PyNET-v2~\cite{ignatov2022pynetv2mobileefficientondevice} and MicroISP~\cite{ignatov2022microispprocessing32mpphotos} have been developed for real-time execution on devices constrained by their hardware resources.  
The advancement in the field has been encouraged by competitions, including the Learned Smartphone ISP Challenge, part of Mobile AI (MAI) workshops in conjunction with CVPR 2021~\cite{ignatov2021learnedsmartphoneispmobile} and CVPR 2022~\cite{ignatov2022learnedsmartphoneispmobile}.

During the aforementioned challenges, teams were invited to submit their models and address two tracks. The solutions were designed to optimize the trade-off between runtime and fidelity (measured by PSNR) in the first track, while the second track has focus on perceptual quality and is evaluated with MOS. A valuable software used by participants, the AI Benchmark application~\cite{ignatov2018aibenchmarkrunningdeep} provides an environment to test Tensorflow Lite models on Android smartphones, taking advantage of the supported acceleration options. 

\begin{figure}[!t]
    \centering
    \includegraphics[width=0.95\linewidth]{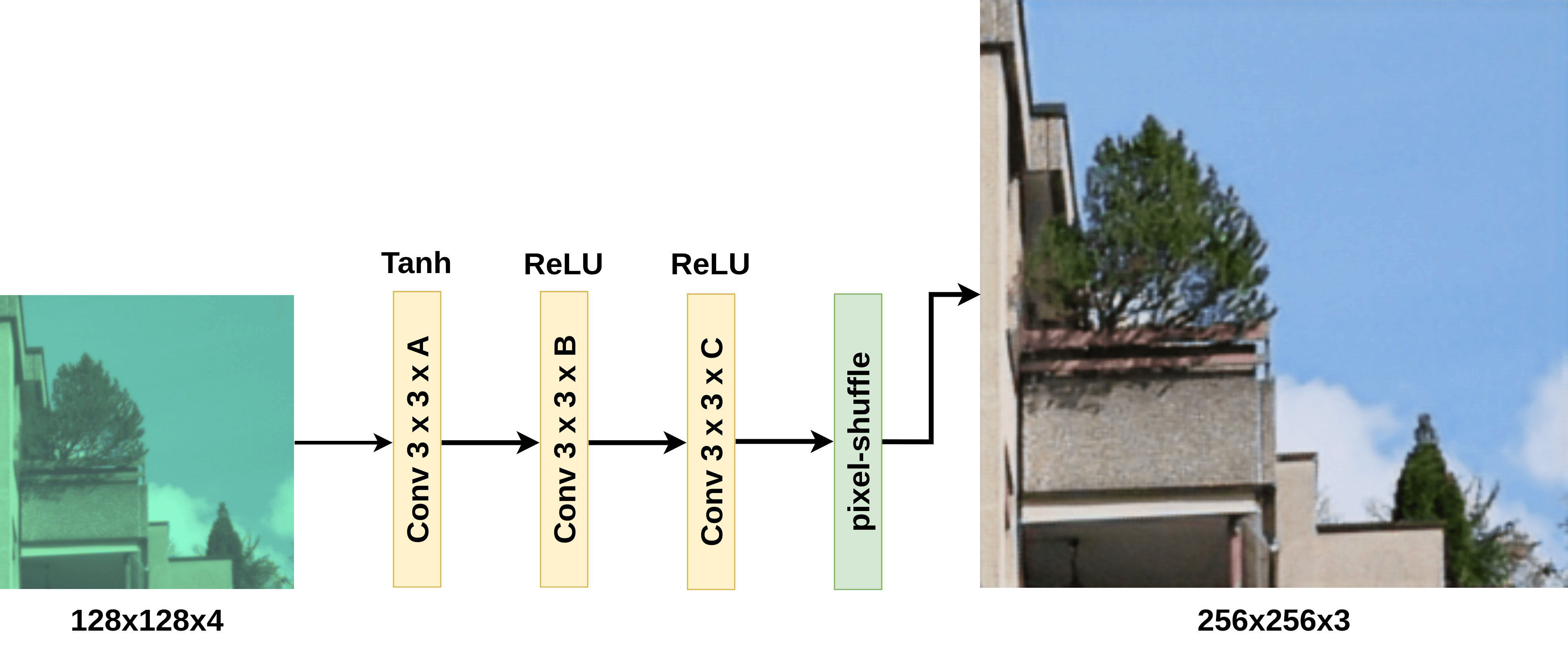}
        \caption{
        Overview of the winning architectures from the MAI 2021 and 2022 challenges. The dh\_isp team \cite{ignatov2021learnedsmartphoneispmobile} uses channel sizes [16, 16, 16], while MiAlgo \cite{ignatov2022learnedsmartphoneispmobile} uses [12, 12, 12]. }
    \label{fig_baseline_arch}
\end{figure}

In both editions, compact networks (\cref{fig_baseline_arch}) with three convolutional layers followed by a pixel-shuffle layer achieved the highest score according to the formula adopted for track one, as they provided the best balance between processing speed and output quality. 

\begin{figure*}[!ht]
    \centering
    \includegraphics[width=\linewidth]{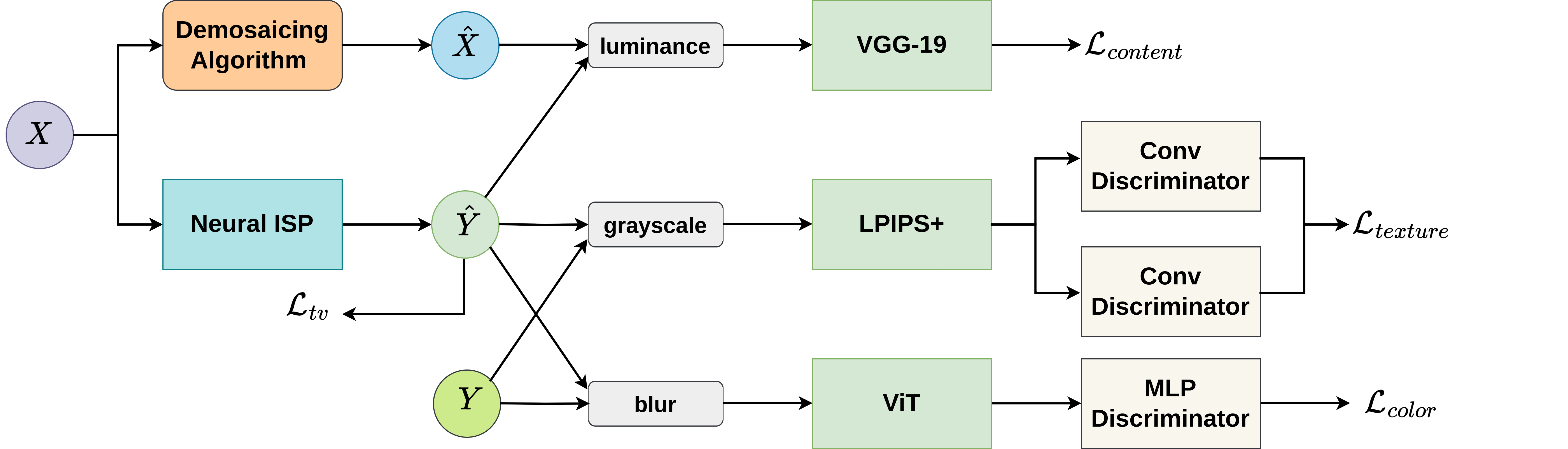}
    \caption{Overview of our proposed unpaired training method.}
    \label{fig:architecture}
\end{figure*}

CSANet~\cite{Hsyu_2021_CVPR} obtained the best results in terms of both PSNR and SSIM in the MAI 2021 challenge. At the core of the network are two double attention modules with skip connections able to learn the spatial and channel dependencies from feature maps. The desired output size of the image is obtained by further use of transpose convolution and depth-to-space layer. LAN~\cite{9856970} is built on this work and increases performance by introducing several improvements to the original architecture. A strided convolutional layer is applied instead of space-to-depth operation at the beginning to improve sharpness and lower GPU latency on smartphones. High-frequency details are better preserved through the implementation of a high-level skip connection via concatenation. Other differences include model pretraining with classical demosaicing, a custom loss function composed of multiple components, and adjustments on the selection of activation functions. Attention mechanisms have also shown notable performance when integrated into U-Net-like architectures that incorporate discrete wavelet transforms (DWT) to emphasize the representation of fine-grained structures, as demonstrated by MW-ISPNet~\cite{ignatov2020aim2020challengelearned} and AWNet~\cite{dai2020awnetattentivewaveletnetwork}.

RMFA-Net~\cite{li2024rmfanetneuralispreal} is a recently proposed network architecture that achieves state-of-the-art (sota) image quality on the Fujifilm UltraISP data set~\cite{ignatov2022microispprocessing32mpphotos}. The architecture consists of an input module composed of two convolutional layers, followed by a stack of RMFA blocks, and an output module implemented as a single convolutional layer. Each RMFA block includes a texture module that processes both high-frequency and low-frequency textures, a tone mapping module based on Retinex Theory~\cite{Land1971LightnessAR}, a spatial attention module, and a channel attention module. Each block also incorporates a skip connection to prevent information loss. In the preprocessing phase, the authors propose a three-channel split and black level subtraction, which play a substantial role in model performance.

Generative Adversarial Networks (GANs) have demonstrated strong capabilities in transferring feature representations across different domains. Reference methods such as CycleGAN~\cite{zhu2020unpairedimagetoimagetranslationusing}, UNIT~\cite{liu2018unsupervisedimagetoimagetranslationnetworks}, and U-GAT-IT~\cite{kim2020ugatitunsupervisedgenerativeattentional} employ cycle consistency constraints, allowing unpaired image-to-image translation. Dedicated discriminators have been used successfully in prior works such as \cite{ignatov2017dslrqualityphotosmobiledevices} and \cite{ignatov2018wespeweaklysupervisedphoto} to guide the learning of color and texture representations in the context of image enhancement. Various GAN objectives have been explored to stabilize training and improve the quality of results, starting from the original adversarial loss~\cite{goodfellow2014generativeadversarialnetworks}, to the Wasserstein loss~\cite{arjovsky2017wassersteingan} with gradient penalty \cite{gulrajani2017improvedtrainingwassersteingans}, the relativistic GAN loss \cite{jolicoeurmartineau2018relativisticdiscriminatorkeyelement}, and more recently the regularized relativistic GAN loss (R3GAN) \cite{huang2025gandeadlonglive}. Researchers commonly design neural networks for ISP learning using multi-term loss functions, where each term targets a specific attribute. Among the frequently used loss functions for capturing pixel-level differences are L1 and L2 losses, along with more robust variants such as Huber and Charbonnier, which are aimed at reducing sensitivity to outliers. SSIM and MS-SSIM are widely used to assess structural similarity based on local image patterns. To improve color fidelity, a common strategy is to first apply Gaussian blur to reduce the influence of textures, followed by measuring color differences in a suitable color space.  DISTS~\cite{Ding_2020} combines deep features with structural information for a more perceptually aligned measure. Perceptual similarity is often measured using feature activations \cite{johnson2016perceptuallossesrealtimestyle} from pretrained networks such as VGG-19 \cite{simonyan2015deepconvolutionalnetworkslargescale}. LPIPS \cite{zhang2018unreasonableeffectivenessdeepfeatures} builds on this idea by computing distances in a deep feature space, fine-tuned to match human perceptual preferences. Our work builds upon several of these definitions, with a central focus on maximizing perceptual image quality.

\section{Proposed Method}
\label{sec:proposed_method}

Although we prioritize fast inference, the training process is not limited by computational cost. Therefore, we incorporate additional networks for adversarial learning and feature extraction (\cref{fig:architecture}). Alongside the unpaired training strategy, we define a method that uses paired data, serving as an upper bound for our unpaired approach.

The color, texture, and content attributes are captured using dedicated loss functions. For each, we utilize feature embeddings extracted from pre-trained networks during loss computation, rather than relying directly on pixel-wise differences of the generated images. Additionally, a total variation (TV) loss is applied to promote spatial smoothness and reduce visual artifacts. A detailed description of each loss function used in our experiments is provided in the following sections.

\subsection{Content Loss}

Following the approach in WESPE~\cite{ignatov2018wespeweaklysupervisedphoto}, structural consistency is enforced by computing the Mean Squared Error (MSE) between the feature maps of the generated image and the corresponding reference. These feature maps are extracted from the relu\_5\_4 layer of the VGG-19 network.

\begin{equation}
  \mathcal{L}_{\text{content}} = \frac{1}{CHW} \sum_{i,j,k} (F_{ijk}^{\text{relu5\_4}}(I_{\text{1}}) - F_{ijk}^{\text{relu5\_4}}(I_{\text{2}}))^2
  \label{eq:content_loss}
\end{equation}

In the paired setting, the generated image and its corresponding ground truth are directly passed into VGG-19. In the unpaired setting, the reference is obtained by applying a specialized demosaicing algorithm to the RAW input. The generated and reference RGB images are then converted to the LAB color space. Only the L (luminance) channel is retained and replicated across all three channels to match the input format required by VGG-19. We denote this loss as $\mathcal{L}_{\text{content (paired)}}$ or $\mathcal{L}_{\text{content (unpaired)}}$, depending on the data access type.

\subsection{Paired Color Loss}

As introduced in DPED~\cite{ignatov2017dslrqualityphotosmobiledevices}, a Gaussian kernel is applied to the images prior to computing the MSE to better quantify the color discrepancies.

\begin{equation}
  \mathcal{L}_{\text{color}} = \frac{1}{N} \sum_{i,j,k} \left( B(I_{\text{1}})_{ijk} - B(I_{\text{2}})_{ijk} \right)^2
  \label{eq:blur_loss}
\end{equation}

\begin{equation}
B(I)_{ijk} = \sum_{m,n} G(m, n) \cdot I_{i, j+m, k+n}
\label{eq:blur_function}
\end{equation}

\begin{equation}
G(m,n) = A \exp \left( -\frac{(m - \mu_x)^2}{2\sigma_x^2} - \frac{(n - \mu_y)^2}{2\sigma_y^2} \right)
\label{eq:gaussian_kernel}
\end{equation}

This approach reduces the influence of fine textures and has been shown to improve contrast and brightness while preserving color fidelity. The resulting loss is referred to as $\mathcal{L}_{\text{color}}$ in the total loss calculation. Moreover, it is tolerant to small pixel misalignments and has been adopted by recently developed sota methods~\cite{li2024rmfanetneuralispreal}.

\subsection{Paired Texture Loss}

We integrate LPIPS+~\cite{chen2023topiqtopdownapproachsemantics} together with DISTS \cite{Ding_2020} as loss components ($\mathcal{L}_{\text{LPIPS+}}$, $\mathcal{L}_{\text{DISTS}}$) responsible for texture and perceptual learning using paired input and ground truth images.

LPIPS (Learned Perceptual Image Patch Similarity) measures perceptual similarity by comparing deep features of two images across multiple layers with fine-tuned weights calibrated to match human visual perception. 

LPIPS+ extends LPIPS by using reference image features as semantic weights in a weighted average pooling scheme. This focuses the metric on important semantic regions, resulting in perceptual quality assessments that better align with human judgments.

DISTS assesses the perceptual quality of an image by combining structure similarity, calculated using normalized correlation between corresponding feature maps, and texture similarity, calculated using normalized similarity between their spatial means, across multiple layers of a modified pre-trained VGG~\cite{simonyan2015deepconvolutionalnetworkslargescale} network.

\begin{figure}[!t]
    \centering
\includegraphics[width=0.85\linewidth]{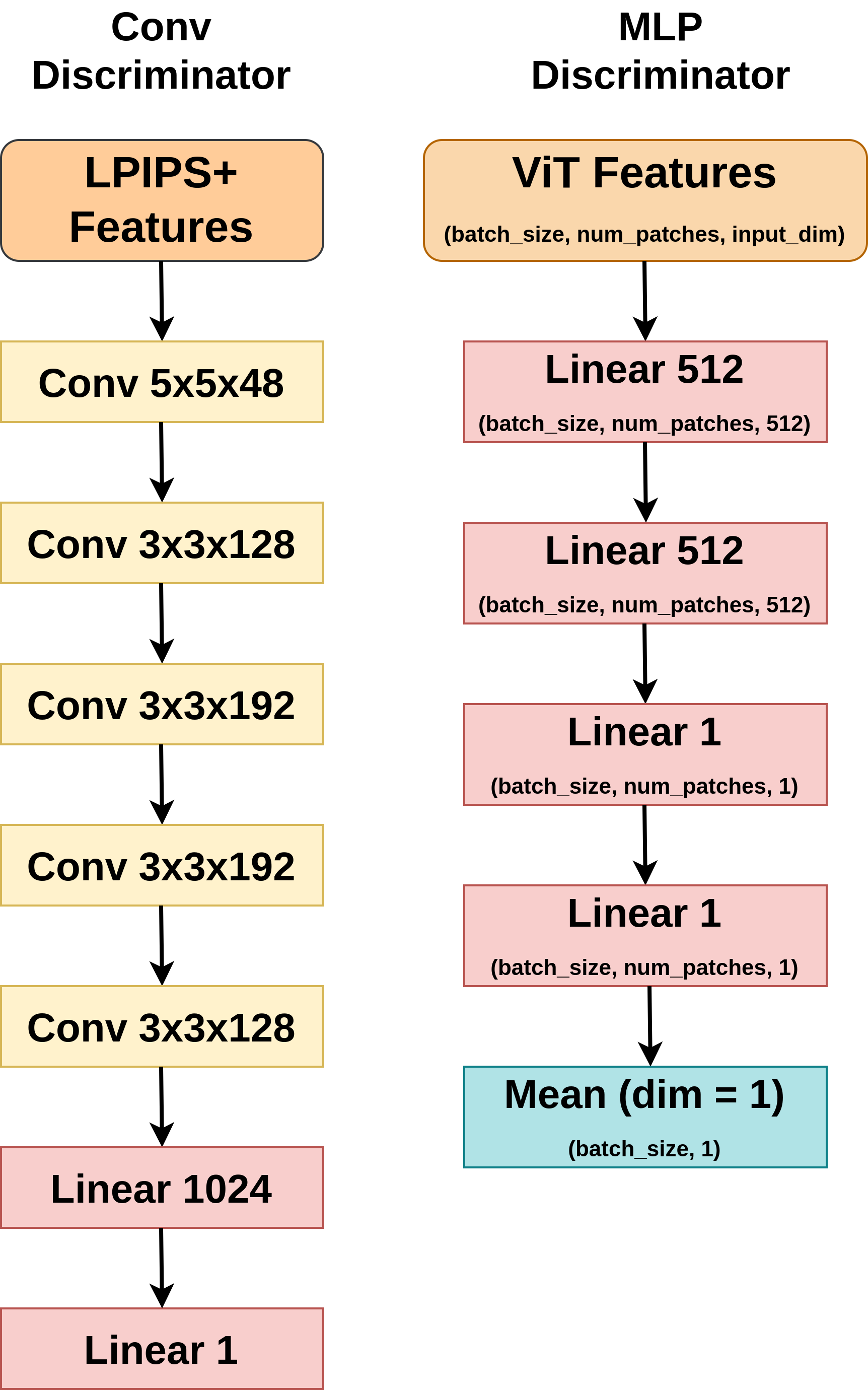}
        \caption{Architecture of the discriminators presented in \cref{fig:architecture}. }
    \vspace{-5mm}
    \label{fig_discriminators}
\end{figure}

\subsection{Relativistic Adversarial Color Loss}

\begin{align}
    R_1 &= \frac{\gamma}{2} \mathbb{E}_{x_r \sim P} 
    \left[ \|\nabla_{x_r} D(x_r)\|^2 \right] \label{eq:R1} \\
    R_2 &= \frac{\gamma}{2} \mathbb{E}_{x_f \sim Q} 
    \left[ \|\nabla_{x_f} D(x_f)\|^2 \right] \label{eq:R2} \\
    L_D &= \mathbb{E}_{x_r, x_f} 
    \left[ \text{f}(- (D(x_r) - D(x_f))) \right] 
     + R_1 + R_2 \label{eq:LD} \\
    L_G &= \mathbb{E}_{x_r, x_f} 
    \left[ \text{f}(- (D(x_f) - D(x_r))) \right] \label{eq:LG}
\end{align}

Unpaired coloring is learned in an adversarial manner. We adopt a relativistic loss \cite{jolicoeurmartineau2018relativisticdiscriminatorkeyelement} with zero-centered gradient penalties \cite{huang2025gandeadlonglive}. The real and generated images are initially fed into a pre-trained ViT-base-patch16-224 model~\cite{dosovitskiy2021imageworth16x16words}. The feature embeddings from the last hidden state of the transformer (excluding the CLS token) are then fed into the color discriminator to predict the realism of the given colors. This adversarial objective constitutes the  $\mathcal{L}_{\text{adv (color)}}$ 
 loss term. The color discriminator consists of a three-layer MLP with ReLU activations, followed by a mean pooling to aggregate the final prediction (\cref{fig_discriminators}).

\subsection{Relativistic Adversarial Texture Loss}

The loss formulas are the same as in the unpaired color loss \cref{eq:LD}, \cref{eq:LG}. Since the LPIPS (and variants) metric is responsible for evaluating perceptual quality, we choose to first convert generated and real images to grayscale and feed them to the LPIPS+ network. 

Then, two discriminators process different levels of LPIPS+ features to evaluate realism from distinct perspectives. In the LPIPS+ architecture, lin0 and lin3 refer to linear layers that process features extracted from different depths of the backbone (AlexNet \cite{alexnet} in IQA-PyTorch \cite{pyiqa} implementation) - lin0 processes features from the first convolutional block (64 channels) capturing low-level details such as edges and sharpness, while lin3 processes features from the fourth block (256 channels) representing more complex patterns, which emphasizes higher-level perceptual quality. The first discriminator receives features from the lin0 layer and contributes with the adversarial loss term $\mathcal{L}_{\text{adv (lin0)}}$ , while the second receives lin3 features and corresponds to $\mathcal{L}_{\text{adv (lin3)}}$. Textural discriminators have a CNN architecture adapted from \cite{ignatov2017dslrqualityphotosmobiledevices} with five convolutional layers followed by two fully connected layers, using Leaky ReLU activations and progressive downsampling of input (\cref{fig_discriminators}).

\subsection{Total Variation Loss}

This loss ($\mathcal{L}_{\text{TV}}$) penalizes differences between adjacent pixels, promoting spatial smoothness in the generated images. It plays an important complementary role to content loss, which effectively preserves high-level structures but often fails to capture fine details. 

\begin{equation}
L_{\text{TV}} = \frac{1}{N} \left( 
\frac{\sum_{i,j} (I_{i+1,j} - I_{i,j})^2}{H(W - 1)C} 
+ \frac{\sum_{i,j} (I_{i,j+1} - I_{i,j})^2}{(H - 1)WC} 
\right)
\end{equation}

If the weight of this loss in the final objective function is too small, unwanted artifacts may occur. Conversely, if its contribution dominates other loss terms, the output images tend to become overly smooth or blurred. Texture-related losses help mitigate this kind of over-smoothing effect.

\subsection{Total Loss Function}

Each training method uses a weighted sum of specific loss terms. The total loss for each stage is defined as follows:

\begin{align}
\mathcal{L}_{\text{paired}} &= \sum_{i} \lambda_i \mathcal{L}_i \label{eq:supervised_loss} \\
\mathcal{L}_{\text{unpaired}} &= \sum_{j} \lambda_j \mathcal{L}_j \label{eq:unsupervised_loss} \\
\mathcal{L}_{\text{pretrain}} &= \sum_{k} \lambda_k \mathcal{L}_k \label{eq:pretrain_loss}
\end{align}

The loss terms used in each stage are as follows:

\begin{itemize}
    \vspace{0.5em}
    \item \textbf{Paired:}
    \[
    \mathcal{L}_i \in \left\{
        \begin{aligned}
        &\mathcal{L}_{\text{content (paired)}}, \mathcal{L}_{\text{LPIPS+}}, \mathcal{L}_{\text{DISTS}},  \\
        &\mathcal{L}_{\text{TV}}, \mathcal{L}_{\text{color}}, \mathcal{L}_{\text{adv (lin0)}}, \mathcal{L}_{\text{adv (lin3)}}
        \end{aligned}
    \right\}
    \]

    \item \textbf{Unpaired:}
    \[
    \mathcal{L}_j \in \left\{
        \begin{aligned}
        &\mathcal{L}_{\text{content (unpaired)}},
        \mathcal{L}_{\text{adv (color)}}, \\
        &\mathcal{L}_{\text{adv (lin0)}}, \mathcal{L}_{\text{adv (lin3)}}, \mathcal{L}_{\text{TV}}
        \end{aligned}
    \right\}
    \]

    \item \textbf{Pretraining:}
    \[
    \mathcal{L}_k \in \left\{ \mathcal{L}_{\text{content (paired)}},
    \mathcal{L}_{\text{MSE}},
    \mathcal{L}_{\text{TV}} \right\}
    \]
\end{itemize}

Each \( \lambda \) denotes the corresponding weight for its associated loss term. The scaling factors are dynamically computed at each training step to ensure that, upon reaching the generator, the gradient norm of each loss component is normalized to 1. This strategy, referred to as Dynamic Loss Adaptation, ensures balanced gradient contributions from all losses during optimization.

When training with adversarial losses, it is important that the model already demonstrates structural consistency and a reasonable level of color reconstruction, as these elements are essential for learning stability. To ensure this, the network is first pre-trained to perform demosaicing on the RAW input with the loss terms specified in \cref{eq:pretrain_loss}.

In our experiments, we consider 3 training scenarios:
\begin{itemize}
    \item Paired data are available, and the formulation from \cref{eq:supervised_loss} is adopted, except for adversarial losses.
    \item Paired data are available, and the formulation described in \cref{eq:supervised_loss} is fully adopted.
    \item The data is unpaired and the training follows the configuration described in \cref{eq:unsupervised_loss}.
\end{itemize}

\section{Experiments}
\label{sec:experiments}

\subsection{Dataset}
\begin{figure}[ht]
\centering
\setlength{\tabcolsep}{1pt}
\scriptsize
\begin{tabular}{ccc}
\textbf{Target Image} & \textbf{Demosaiced RAW} & \textbf{Ours (unpaired)} \\
\includegraphics[width=0.3\linewidth]{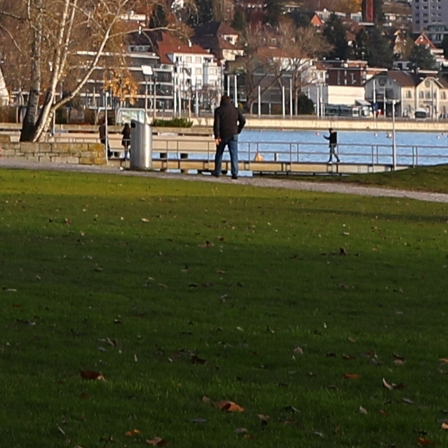} &
\includegraphics[width=0.3\linewidth]{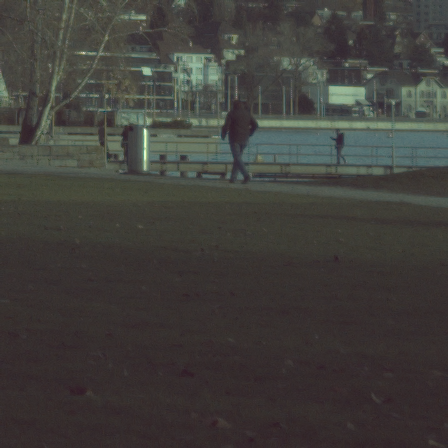} &
\includegraphics[width=0.3\linewidth]{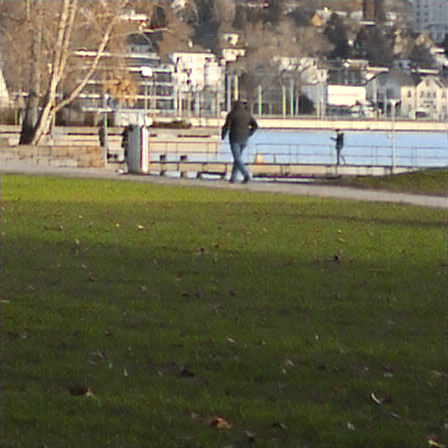} \\
\includegraphics[width=0.3\linewidth]{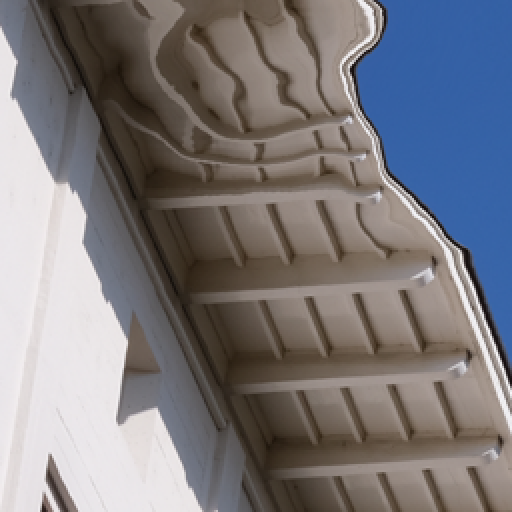} &
\includegraphics[width=0.3\linewidth]{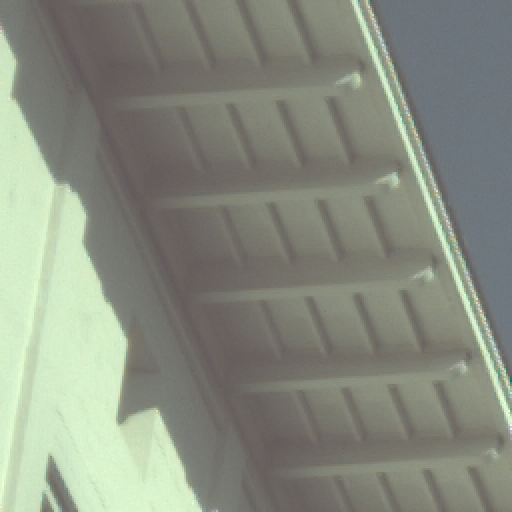} &
\includegraphics[width=0.3\linewidth]{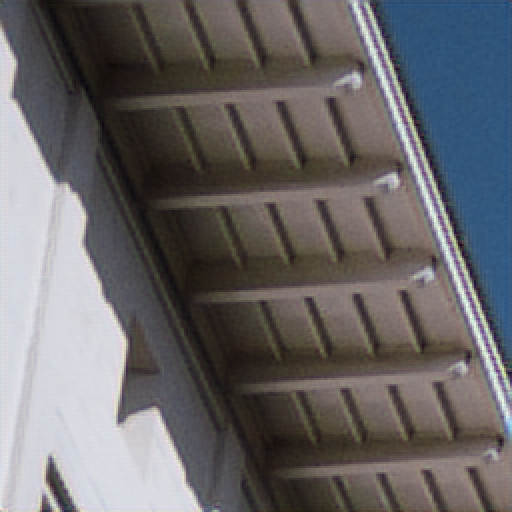} \\
\end{tabular}
\caption{Dataset challenges. The first row shows images from the ZRR dataset (training subset)~\cite{ignatov2020replacingmobilecameraisp}, which include dynamic elements and slight viewpoint misalignments. The second row shows a Fujifilm UltraISP~\cite{ignatov2022microispprocessing32mpphotos} training sample with noticeable warping caused by the alignment algorithm.}
\label{fig:example_misalignment}
\end{figure}

Our method is evaluated on the ZRR~\cite{ignatov2020replacingmobilecameraisp} and Fujifilm UltraISP~\cite{ignatov2022microispprocessing32mpphotos} datasets to demonstrate its generalization across differing data distributions. A key advantage of our method is that since it does not require paired data at the content level, it is robust by design to various sources of misalignment (\cref{fig:example_misalignment}).

Each RAW training image from the ZRR dataset~\cite{ignatov2020replacingmobilecameraisp} is captured using a 12.3 MP Sony Exmor IMX380 Bayer sensor and paired with a corresponding image generated by a high-end Canon 5D Mark IV camera. For global alignment, SIFT keypoints \cite{sift} and RANSAC \cite{ransac} were used, followed by patch extraction (448×448) using a sliding window. To further refine the alignment, the patch positions were adjusted to maximize cross-correlation, resulting in 48K aligned RAW–RGB samples. From this set, 1.2K pairs were reserved for testing, the remainder being used for training and validation. This data set is entirely available to the public.

\begin{figure*}[t!]
\centering
\renewcommand{\arraystretch}{1.1}
\setlength{\tabcolsep}{1pt}
\scriptsize
\resizebox{\linewidth}{!}{
\begin{tabular}{@{}c|ccc|c@{}}
 & \multicolumn{3}{c|}{\textbf{Paired Training}} & \textbf{Unpaired Training} \\
\textbf{Ground Truth} & \textbf{LAN}~\cite{9856970} & \textbf{Ours w/o adv. losses} & \textbf{Ours } & \textbf{Ours } \\
\includegraphics[width=0.185\linewidth]{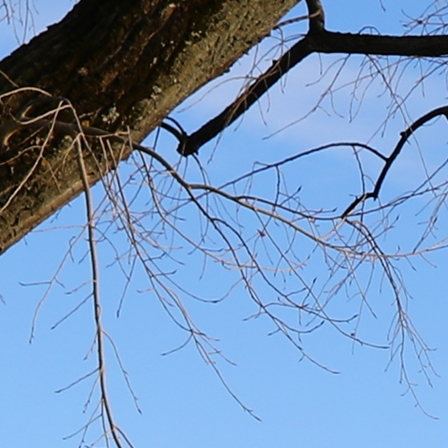} &
\includegraphics[width=0.185\linewidth]{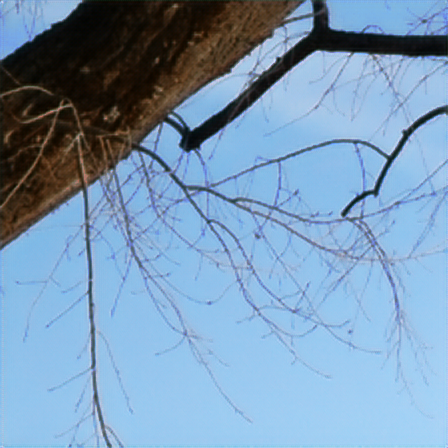} &
\includegraphics[width=0.185\linewidth]{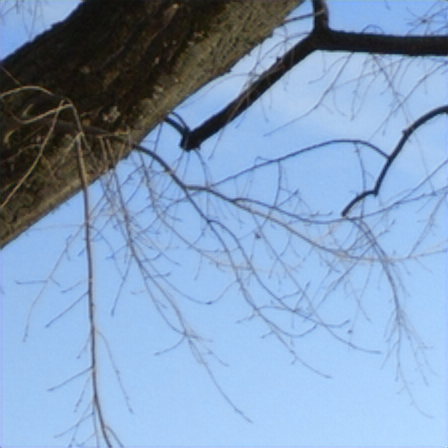} &
\includegraphics[width=0.185\linewidth]{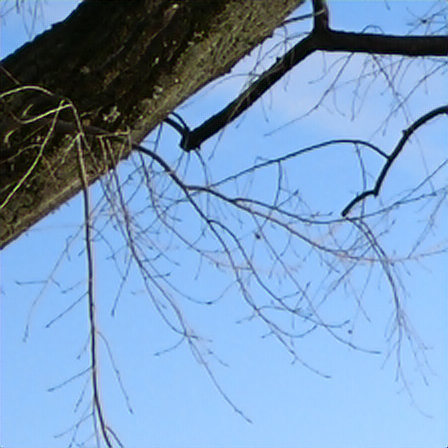} &
\includegraphics[width=0.185\linewidth]{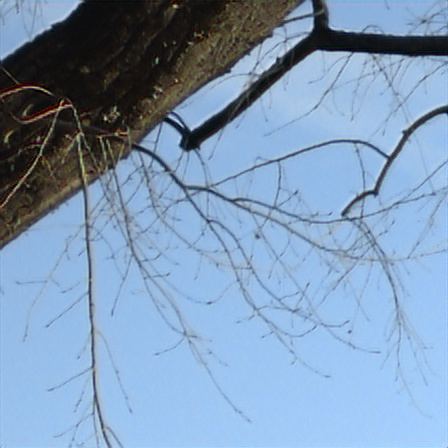} \\
\includegraphics[width=0.185\linewidth]{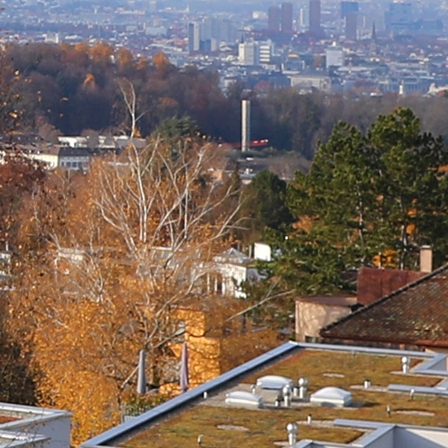} &
\includegraphics[width=0.185\linewidth]{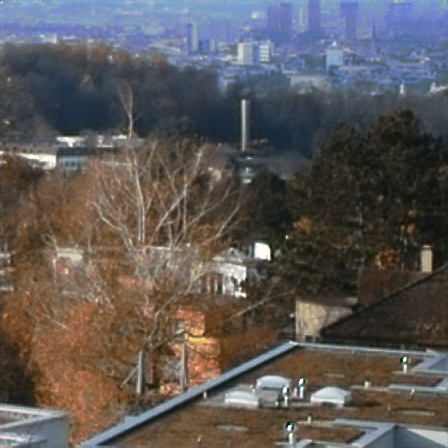} &
\includegraphics[width=0.185\linewidth]{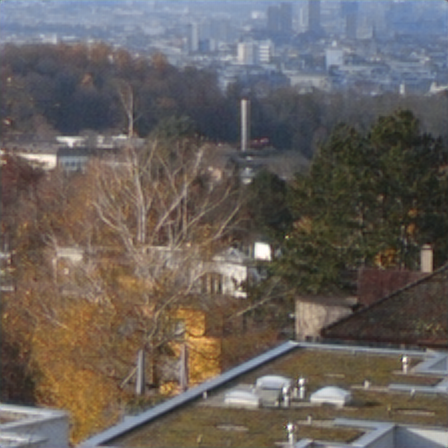} &
\includegraphics[width=0.185\linewidth]{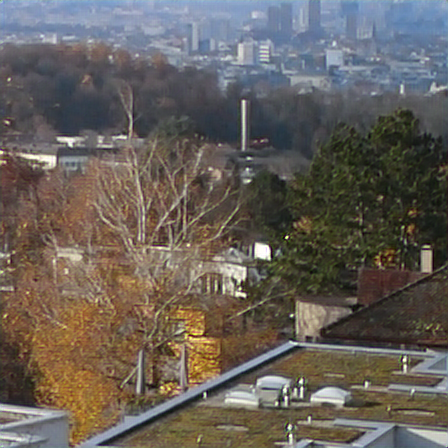} &
\includegraphics[width=0.185\linewidth]{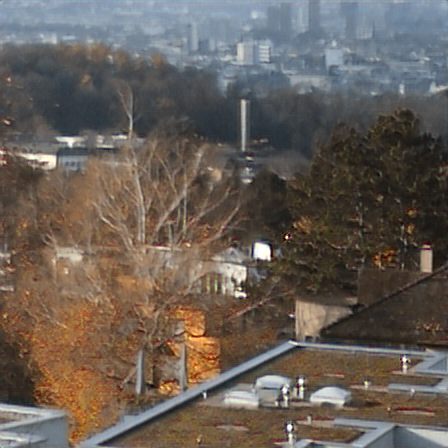} \\
\includegraphics[width=0.185\linewidth]{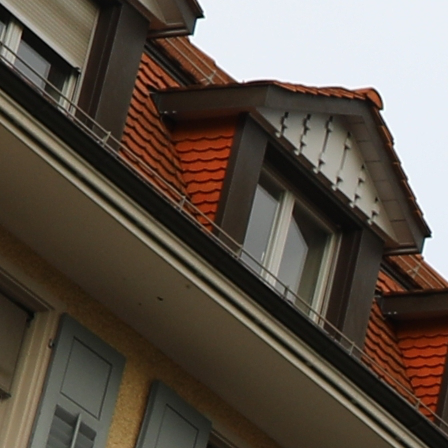} &
\includegraphics[width=0.185\linewidth]{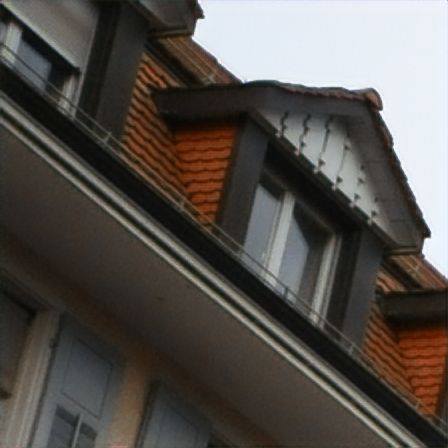} &
\includegraphics[width=0.185\linewidth]{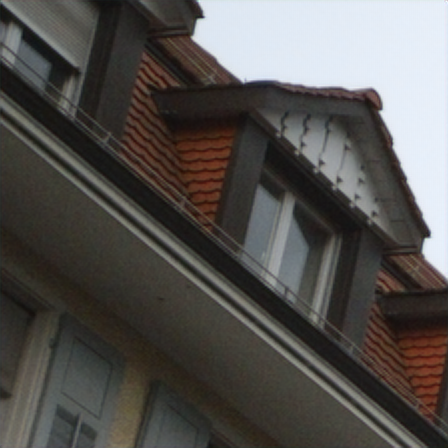} &
\includegraphics[width=0.185\linewidth]{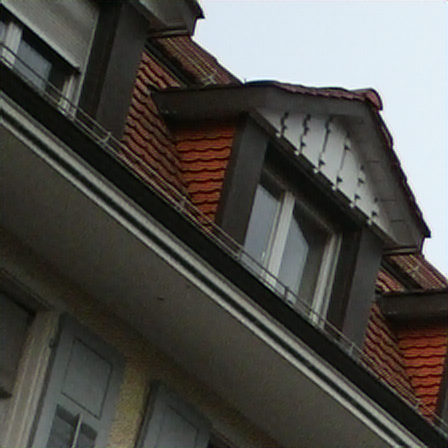} &
\includegraphics[width=0.185\linewidth]{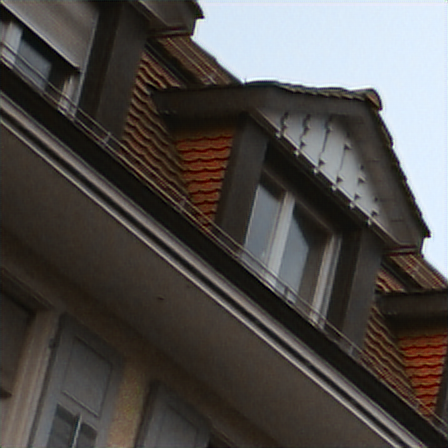} \\
\includegraphics[width=0.185\linewidth]{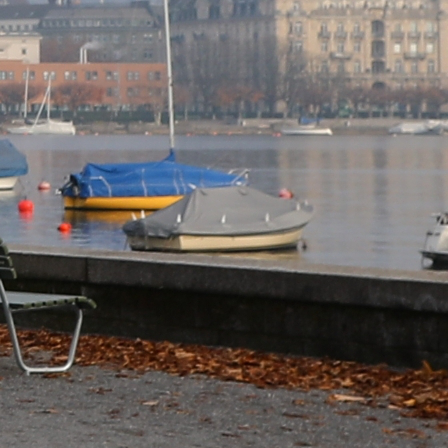} &
\includegraphics[width=0.185\linewidth]{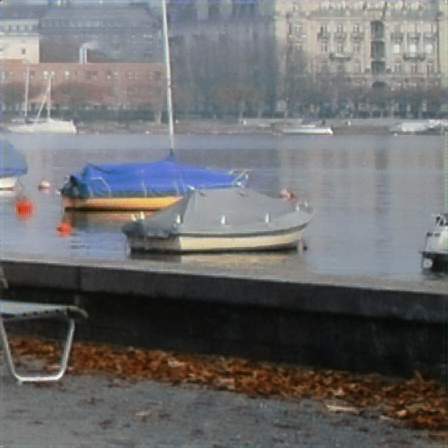} &
\includegraphics[width=0.185\linewidth]{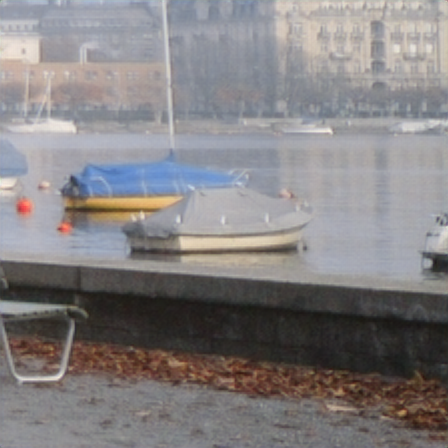} &
\includegraphics[width=0.185\linewidth]{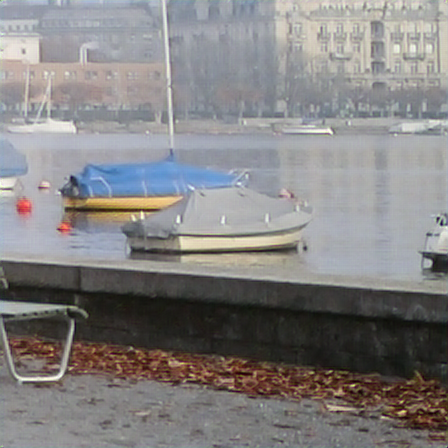} &
\includegraphics[width=0.185\linewidth]{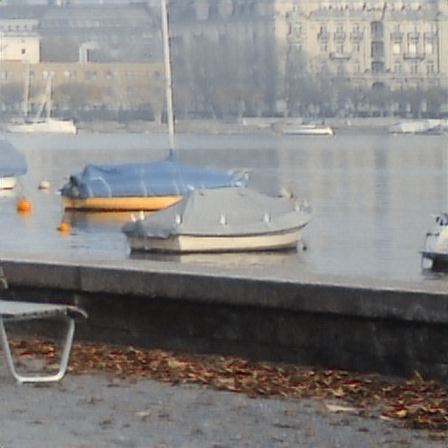} \\

\end{tabular}
}
\caption{
  Visual comparisons of outputs and target images on ZRR dataset (test subset)~\cite{ignatov2020replacingmobilecameraisp}. Last three columns show visual results of Efficient ISP trained under different data access settings.
}
\vspace{-2mm}
\label{fig:enhancedexamples}
\end{figure*}

In Fujifilm UltraISP~\cite{ignatov2022microispprocessing32mpphotos}, the authors used a Sony IMX586 Quad Bayer camera sensor and a Fujifilm GFX100 DSLR to acquire visual data. To enhance alignment with the demosaiced input, target images were processed using PDC-Net~\cite{truong2021learningaccuratedensecorrespondences}, followed by the extraction of 256×256 pixel patches. Only training pairs and raw validation patches were publicly released. Participants could upload their output through the contest platform and receive PSNR and SSIM scores for the official validation set.

When developing locally, we first removed 17.6\% of the training samples from the Fujifilm UltraISP dataset due to small misalignments and then split the remaining data so that 1,024 images were used for validation and another 1,024 for testing. For evaluation on the ZRR dataset, we randomly sampled 1,024 images from the training set to create a validation set.

\begin{table*}[t]
  \centering
  \resizebox{\linewidth}{!}{
  \begin{tabular}{@{}l l l c c c c c c@{}}
    \toprule
    \textbf{Data Access} & \textbf{Method} & \textbf{Backbone} & \textbf{Params} $\downarrow$ & \textbf{Latency (ms)} $\downarrow$ & \textbf{PSNR} $\uparrow$ & \textbf{SSIM} $\uparrow$ & \textbf{MS-SSIM} $\uparrow$ & \textbf{LPIPS} $\downarrow$ \\
    \midrule
    \multirow{4}{*}{\textbf{Paired}} 
    & ours w/o adv.losses & Efficient ISP & 3K & 32.9 & 19.667 & 0.699 & 0.831 & 0.240 \\
    & ours & Efficient ISP & 3K & 32.9 & 19.692 & 0.699 & 0.837 & \textbf{0.198} \\
    & original~\cite{9856970} & LAN & 46K & 241 & \textbf{20.403} & 0.697 & \textbf{0.843} & 0.213 \\
    & original~\cite{dong2015imagesuperresolutionusingdeep} & SRCNN & 25K & 1013 & 19.833 & 0.693 & 0.808 & 0.370 \\
    \midrule
    \textbf{Unpaired} & ours & Efficient ISP & 3K & 32.9 & 19.448 & \textbf{0.700} & 0.832 & 0.239 \\
    \bottomrule
  \end{tabular}}
  \caption{Evaluation on ZRR test data~\cite{ignatov2020replacingmobilecameraisp}. All models were trained on ZRR. Inference time is measured on mobile GPU on a Full HD (1920 × 1088) image.}
  \label{tab:zrr_comparison}
\end{table*}

\begin{table*}[ht]
  \centering
  \resizebox{\linewidth}{!}{
    \begin{tabular}{@{}l l l | c c c c @{\hskip 6pt} | c c c c @{\hskip 6pt} | c c  @{}}
      \toprule
      \textbf{Data}&\multicolumn{2}{c|}{} & \multicolumn{4}{c|}{\textbf{Local Validation}} & \multicolumn{4}{c|}{\textbf{Local Test}} & \multicolumn{2}{c}{\textbf{Competition Valid.}} \\
      \textbf{Access} & \textbf{Method} & \textbf{Backbone} & \textbf{PSNR} $\uparrow$  & \textbf{SSIM} $\uparrow$ & \textbf{MS-SSIM} $\uparrow$ & \textbf{LPIPS} $\downarrow$ 
            & \textbf{PSNR} $\uparrow$  & \textbf{SSIM} $\uparrow$ & \textbf{MS-SSIM} $\uparrow$ & \textbf{LPIPS} $\downarrow$  
            & \textbf{PSNR} $\uparrow$  & \textbf{SSIM} $\uparrow$  \\
      \midrule
        \multirow{6}{*}{\textbf{Paired}} 
        & \multirow{3}{*}{\shortstack{ours w/o\\adv.losses}} & RMFA-Net tiny & 23.6253 & 0.8256  & 0.9354 & 0.1597  & 23.7468 & 0.8321  & 0.9360 & 0.1545  & 23.52 & 0.82   \\
        & & Efficient ISP & 23.1083 & 0.8214  &  0.9382 & 0.1902  & 23.2692 & 0.8273  & 0.9374 & 0.1851 & 23.24 & 0.81 \\
        & & Robust ISP & 23.0679 & 0.8236  & 0.9414 & 0.1960  & 23.1823 & 0.8286  &0.9402 & 0.1930  & 23.17 & 0.81 \\
        \cmidrule(lr){2-13}
        & \multirow{3}{*}{ours} & RMFA-Net tiny & 24.4667 & 0.8653 & 0.9573 & 0.1147  & 24.5764 & 0.8698 & 0.9572 & 0.1121  & 24.27 & 0.85 \\
        & & Efficient ISP &  23.5346 & 0.8367  & 0.9502 & 0.1490   & 23.6477 &  0.8401  & 0.9498 &  0.1465 & 23.65 & 0.83   \\
        & & Robust ISP & 23.3905 &  0.8381   & 0.9489 &  0.1557  & 23.4914 & 0.8411 & 0.9482 & 0.1547  & 23.52 & 0.83\\  
      \midrule
        \multirow{3}{*}{\textbf{Unpaired}} 
        & \multirow{3}{*}{ours} & RMFA-Net tiny & 22.8727 & 0.8399  & 0.9455 &  0.1593  & 22.7739 & 0.8451 & 0.9452 & 0.1586  & 22.75 & 0.83 \\
        & & Efficient ISP & 23.0275 & 0.8259   & 0.9457 &  0.1635  & 23.0711 & 0.8300   & 0.9449 & 0.1614   & 23.10 & 0.81 \\
        & & Robust ISP & 22.7885 & 0.8259   &  0.9446 & 0.1785    & 22.9401 & 0.8309 & 0.9445 & 0.1751   & 22.97 & 0.81 \\
      \bottomrule
    \end{tabular}}
  \caption{Quantitative results on the Fujifilm UltraISP dataset~\cite{ignatov2022microispprocessing32mpphotos} and on the Mobile AI 2025 competition validation data~\cite{MAI2025learnedISP}.}
  \label{tab:custom_val_fujifilm}
\end{table*}

\subsection{Ablation Study}

\begin{table}[t]
  \centering
    \resizebox{\linewidth}{!}{
    \begin{tabular}{@{}l c c c c c@{}}
      \toprule
      \textbf{Demosaicing} & $Beta_{1}$ & \textbf{PSNR} $\uparrow$ & \textbf{SSIM} $\uparrow$ & \textbf{MS-SSIM} $\uparrow$ & \textbf{LPIPS} $\downarrow$ \\
      \midrule
      \multirow{2}{*}{\shortstack{OpenCV \\ (BG2RGB)}} 
      & 0.0 & 19.4509 & 0.6925 & 0.8294 & 0.2351 \\
      & 0.5 & \textbf{19.5277} & \textbf{0.7009} & \textbf{0.8322} & 0.2396 \\
      \midrule
      \multirow{2}{*}{\shortstack{Menon~\cite{4032820} \\ 2007}} 
      & 0.0 & 19.4545 & 0.6916 & 0.8301 & 0.2336 \\
      & 0.5 & 19.4809 & 0.6907 & 0.8290 & \textbf{0.2328} \\
      \bottomrule
    \end{tabular}}
  \caption{
  Results reported on ZRR test set~\cite{ignatov2020replacingmobilecameraisp}. Efficient ISP was trained under our unpaired setting with different demosaicing algorithms and momentum values. }
  \label{tab:zrr_settings}
\end{table}

\begin{table}[t]
    \centering
    \resizebox{\linewidth}{!}{
    \begin{tabular}{@{}l c c c c@{}}
      \toprule
      \textbf{Layer} & \textbf{PSNR} $\uparrow$ & \textbf{SSIM} $\uparrow$ & \textbf{MS-SSIM} $\uparrow$ & \textbf{LPIPS} $\downarrow$ \\
      \midrule
        lin0  & \textbf{19.1230} & \textbf{0.6661} & 0.8139 & 0.2841 \\
        lin1  & 18.6935 & 0.6639 & \textbf{0.8152} & 0.2920 \\
        lin2  & 18.3013 & 0.6514 & 0.8065 & 0.2950 \\
        lin3  & 18.2038 & 0.6539 & 0.8090 & \textbf{0.2838} \\
        lin4  & 17.9374 & 0.6474 & 0.7962 & 0.3121 \\
      \bottomrule
    \end{tabular}}
  \caption{Performance comparison of Efficient ISP trained with unpaired data from the ZRR~\cite{ignatov2020replacingmobilecameraisp} dataset. Results are reported on ZRR test set using a single discriminator conditioned on different LPIPS+~\cite{chen2023topiqtopdownapproachsemantics} feature map layers.}
  \label{tab:texture_ablation}
\end{table}

To effectively guide texture learning through a GAN-based loss, the discriminator should receive information that emphasizes features relevant to the target objective, while suppressing non-essential ones. Previous works have addressed this by converting the image to grayscale before passing it to the discriminator, removing color information, and allowing the model to focus more on structural and textural details. However, texture is strongly correlated with perceptual quality, which is often evaluated using the LPIPS~\cite{zhang2018unreasonableeffectivenessdeepfeatures} measure. Since LPIPS computes a weighted difference of deep features across multiple layers, feeding such feature representations directly into the discriminator can promote better LPIPS performance. As a result, it leads to improved texture reconstruction and perceptual realism. Different layers capture progressively more abstract feature maps, ranging from low-level information to high-level representations as the network goes deeper. Among the tested configurations, the set-up that uses one discriminator to learn features from lin0 and another from lin3 provides the best results. This outcome is expected, as both discriminators individually outperformed the others in previous experiments on the perceptual score (\cref{tab:texture_ablation}). The lin0-based discriminator plays a key role in counteracting the potential over-smoothing introduced by the total variation loss by emphasizing high-frequency information, including fine details and noise. In contrast, the lin3-based discriminator reduces unwanted noise without compromising the structural fidelity enhanced by the lin0-based one.

Using blurred versions of the images, normalized and fed into the convolutional discriminator, has proven effective for learning color. Although with this approach, we achieved comparable performance, we observed that passing the blurred image through a pre-trained network and feeding the resulting feature maps to the discriminator led to significantly faster convergence, more stable training, and reduced variation caused by updates. We opted for a Vision Transformer architecture as it demonstrated better learning stability compared to other options such as VGG or AlexNet.

In general, the desired balance required to make the GAN training work requires a careful choice of hyperparameters and sometimes additional empirical adjustments. An important hyperparameter is momentum (Adam $Beta_{1}$), whose value was directly related to training success, as highlighted in \cite{huang2025gandeadlonglive}. To address stability and generalization in the unpaired approach, we performed experiments with different values for $Beta_{1}$, as well as different demosaicing algorithms, \ie Menon 2007~\cite{4032820} and the OpenCV built-in demosaicing method (BG2RGB). The results in  \cref{tab:zrr_settings} show consistent performance across settings, indicating robustness in this regard.

\paragraph{Backbones}
We adopt the network architecture proposed by the winner of \cite{ignatov2022learnedsmartphoneispmobile}, hereafter referred to as Efficient ISP. It consists of only three convolutional layers with 3×3 kernels and 12 channels each, followed by a pixel-shuffle layer. The first activation function is Tanh, while ReLU is used in the subsequent layers. Despite its simplicity, the model demonstrated good performance and notable computational efficiency in the competition.

We evaluated the performance of Efficient ISP, trained with paired and unpaired data, in comparison to two locally trained sota models: LAN~\cite{9856970}, using original source code provided by the authors and a custom implementation of SRCNN~\cite{dong2015imagesuperresolutionusingdeep}.  We also explored alternative channel configurations and proposed a second backbone with 16, 4, and 12 channels, called a Robust ISP. The latter has been chosen because it is faster, has fewer parameters, and provides a higher competition score \cite{ignatov2022learnedsmartphoneispmobile} when measured locally. The tiny version of RMFA-Net~\cite{li2024rmfanetneuralispreal} is our third backbone. It is designed to run on smartphones and the authors reported the sota performance on MAI 2022 data set of 24.549 dB in PSNR, which is consistent with our results in paired setting.

As shown in \cref{tab:zrr_comparison}, the models trained with our method perform well in structural metrics, achieving particularly favorable LPIPS scores, the main objective in our case. In the challenge dataset, the unpaired method generally outperforms the paired variant without GAN-based losses, both structurally and perceptually. Besides the contribution of textural component, one factor implied is that the unpaired method's content loss remains unaffected by potential misalignments. The results are consistent across all three splits, indicating robust generalization. 

It should be mentioned that the winning model of the 2022 edition of the MAI Learned ISP Challenge got 23.33 dB PSNR~\cite{ignatov2022learnedsmartphoneispmobile} in the final ranking. The same network (Efficient ISP), trained with our unpaired strategy, obtained 23.10 dB PSNR on the official validation set and above 23 dB on the other data partitions we used for evaluation~(\cref{tab:custom_val_fujifilm}).

Furthermore, the texture component demonstrated a significant contribution to perceptual quality (LPIPS score) when multiple models were trained with access to paired data (\cref{tab:custom_val_fujifilm}). This effect is consistent in the ZRR dataset, as can be observed in quantitative (\cref{tab:zrr_comparison}) and qualitative (\cref{fig:enhancedexamples}) results.

\subsection{Implementation Details}

The models were trained using Adam optimizer with a batch size of 32. Since the generator deals with a more complex task with its lightweight architecture and receives feedback from multiple sources, the discriminators learning process needs to be slowed down through reduced learning rates and appropriate update ratios. These hyperparameters are required to be fine-tuned for each learnable ISP, depending on the complexity of the network.

Efficient and Robust ISPs used a learning rate of $5 \cdot 10^{-4}$ and their discriminators were trained with a learning rate of $10^{-5}$ at every 10th step. The tiny RMFA-Net used a learning rate of $10^{-4}$ and was trained in a ratio of 4:1 with discriminators using a learning rate of $5 \cdot 10^{-5}$.

We performed the training on multiple cloud virtual machines, each configured with an NVIDIA RTX 4090 GPU. The code was implemented in Pytorch framework and uses the support of IQA-PyTorch Toolbox ~\cite{pyiqa}. 
\section{Conclusion}

In this work, we introduced a new method for training a learnable ISP capable of running on mobile devices without the restriction of paired data. With the same backbone architecture as the 2022 MAI challenge winner, our unpaired data training method achieves a PSNR score only 0.3 dB lower than the original approach that relied on paired images \cite{ignatov2022learnedsmartphoneispmobile}. With the help of discriminators that receive perceptually relevant feature maps from pre-trained networks, the neural ISP is guided to focus on fine details and textures that enhance perceptual quality. When integrated in a paired setting, the adversarial component of texture leads to even greater visual fidelity.

For the paired approach, further improvements in color accuracy and tone mapping can be achieved by integrating NILUT \cite{conde2023nilutconditionalneuralimplicit} as a preprocessing step. To address the challenges in the unpaired training setting, future work will focus on improving training performance through adaptive hyperparameter selection and reducing the fidelity gap, particularly with respect to PSNR, between the results obtained from training with unpaired data and those obtained using paired data.

\label{sec:conclusion}

\section*{Acknowledgments}
This work was partially supported by the Humboldt Foundation.

{
    \small
    \bibliographystyle{ieeenat_fullname}
    \bibliography{main}

\begin{thebibliography}{41}
\providecommand{\natexlab}[1]{#1}
\providecommand{\url}[1]{\texttt{#1}}
\expandafter\ifx\csname urlstyle\endcsname\relax
  \providecommand{\doi}[1]{doi: #1}\else
  \providecommand{\doi}{doi: \begingroup \urlstyle{rm}\Url}\fi

\bibitem[Arjovsky et~al.(2017)Arjovsky, Chintala, and Bottou]{arjovsky2017wassersteingan}
Mart{\'{\i}}n Arjovsky, Soumith Chintala, and L{\'{e}}on Bottou.
\newblock Wasserstein generative adversarial networks.
\newblock In \emph{Proceedings of the 34th International Conference on Machine Learning (ICML)}, pages 214--223. PMLR, 2017.

\bibitem[Chen and Mo(2022)]{pyiqa}
Chaofeng Chen and Jiadi Mo.
\newblock {IQA-PyTorch}: Pytorch toolbox for image quality assessment.
\newblock [Online]. Available: \url{https://github.com/chaofengc/IQA-PyTorch}, 2022.

\bibitem[Chen et~al.(2024)Chen, Mo, Hou, Wu, Liao, Sun, Yan, and Lin]{chen2023topiqtopdownapproachsemantics}
Chaofeng Chen, Jiadi Mo, Jingwen Hou, Haoning Wu, Liang Liao, Wenxiu Sun, Qiong Yan, and Weisi Lin.
\newblock Topiq: A top-down approach from semantics to distortions for image quality assessment.
\newblock \emph{IEEE Transactions on Image Processing}, 2024.

\bibitem[Conde et~al.(2024)Conde, Vazquez-Corral, Brown, and Timofte]{conde2023nilutconditionalneuralimplicit}
Marcos~V Conde, Javier Vazquez-Corral, Michael~S Brown, and Radu Timofte.
\newblock {NILUT}: Conditional neural implicit 3d lookup tables for image enhancement.
\newblock In \emph{Proceedings of the AAAI Conference on Artificial Intelligence}, pages 1371--1379, 2024.

\bibitem[Dai et~al.(2020)Dai, Liu, Li, and Chen]{dai2020awnetattentivewaveletnetwork}
Linhui Dai, Xiaohong Liu, Chengqi Li, and Jun Chen.
\newblock Awnet: Attentive wavelet network for image isp.
\newblock In \emph{Computer Vision--ECCV 2020 Workshops: Glasgow, UK, August 23--28, 2020, Proceedings, Part III 16}, pages 185--201. Springer, 2020.

\bibitem[Ding et~al.(2020)Ding, Ma, Wang, and Simoncelli]{Ding_2020}
Keyan Ding, Kede Ma, Shiqi Wang, and Eero~P. Simoncelli.
\newblock Image quality assessment: Unifying structure and texture similarity.
\newblock \emph{IEEE Transactions on Pattern Analysis and Machine Intelligence}, page 1–1, 2020.

\bibitem[Dong et~al.(2015)Dong, Loy, He, and Tang]{dong2015imagesuperresolutionusingdeep}
Chao Dong, Chen~Change Loy, Kaiming He, and Xiaoou Tang.
\newblock Image super-resolution using deep convolutional networks.
\newblock \emph{IEEE transactions on pattern analysis and machine intelligence}, 38\penalty0 (2):\penalty0 295--307, 2015.

\bibitem[Dosovitskiy et~al.(2021)Dosovitskiy, Beyer, Kolesnikov, Weissenborn, Zhai, Unterthiner, Dehghani, Minderer, Heigold, Gelly, Uszkoreit, and Houlsby]{dosovitskiy2021imageworth16x16words}
Alexey Dosovitskiy, Lucas Beyer, Alexander Kolesnikov, Dirk Weissenborn, Xiaohua Zhai, Thomas Unterthiner, Mostafa Dehghani, Matthias Minderer, Georg Heigold, Sylvain Gelly, Jakob Uszkoreit, and Neil Houlsby.
\newblock An image is worth 16x16 words: Transformers for image recognition at scale.
\newblock In \emph{Proceedings of the 9th International Conference on Learning Representations (ICLR)}. OpenReview.net, 2021.

\bibitem[Goodfellow et~al.(2020)Goodfellow, Pouget-Abadie, Mirza, Xu, Warde-Farley, Ozair, Courville, and Bengio]{goodfellow2014generativeadversarialnetworks}
Ian Goodfellow, Jean Pouget-Abadie, Mehdi Mirza, Bing Xu, David Warde-Farley, Sherjil Ozair, Aaron Courville, and Yoshua Bengio.
\newblock Generative adversarial networks.
\newblock \emph{Communications of the ACM}, 63\penalty0 (11):\penalty0 139--144, 2020.

\bibitem[Gulrajani et~al.(2017)Gulrajani, Ahmed, Arjovsky, Dumoulin, and Courville]{gulrajani2017improvedtrainingwassersteingans}
Ishaan Gulrajani, Faruk Ahmed, Martin Arjovsky, Vincent Dumoulin, and Aaron~C Courville.
\newblock Improved training of wasserstein gans.
\newblock \emph{Advances in neural information processing systems}, 30, 2017.

\bibitem[Hsyu et~al.(2021)Hsyu, Liu, Chen, Chen, and Tsai]{Hsyu_2021_CVPR}
Ming-Chun Hsyu, Chih-Wei Liu, Chao-Hung Chen, Chao-Wei Chen, and Wen-Chia Tsai.
\newblock Csanet: High speed channel spatial attention network for mobile isp.
\newblock In \emph{Proceedings of the IEEE/CVF Conference on Computer Vision and Pattern Recognition (CVPR) Workshops}, pages 2486--2493, 2021.

\bibitem[Huang et~al.(2024)Huang, Gokaslan, Kuleshov, and Tompkin]{huang2025gandeadlonglive}
Nick Huang, Aaron Gokaslan, Volodymyr Kuleshov, and James Tompkin.
\newblock The gan is dead; long live the gan! a modern gan baseline.
\newblock \emph{Advances in Neural Information Processing Systems}, 37:\penalty0 44177--44215, 2024.

\bibitem[Ignatov et~al.(2017)Ignatov, Kobyshev, Timofte, Vanhoey, and Van~Gool]{ignatov2017dslrqualityphotosmobiledevices}
Andrey Ignatov, Nikolay Kobyshev, Radu Timofte, Kenneth Vanhoey, and Luc Van~Gool.
\newblock Dslr-quality photos on mobile devices with deep convolutional networks.
\newblock In \emph{Proceedings of the IEEE international conference on computer vision}, pages 3277--3285, 2017.

\bibitem[Ignatov et~al.(2018{\natexlab{a}})Ignatov, Kobyshev, Timofte, Vanhoey, and Van~Gool]{ignatov2018wespeweaklysupervisedphoto}
Andrey Ignatov, Nikolay Kobyshev, Radu Timofte, Kenneth Vanhoey, and Luc Van~Gool.
\newblock Wespe: weakly supervised photo enhancer for digital cameras.
\newblock In \emph{Proceedings of the IEEE conference on computer vision and pattern recognition workshops}, pages 691--700, 2018{\natexlab{a}}.

\bibitem[Ignatov et~al.(2018{\natexlab{b}})Ignatov, Timofte, Chou, Wang, Wu, Hartley, and Van~Gool]{ignatov2018aibenchmarkrunningdeep}
Andrey Ignatov, Radu Timofte, William Chou, Ke Wang, Max Wu, Tim Hartley, and Luc Van~Gool.
\newblock Ai benchmark: Running deep neural networks on android smartphones.
\newblock In \emph{Proceedings of the European Conference on Computer Vision (ECCV) Workshops}, pages 0--0, 2018{\natexlab{b}}.

\bibitem[Ignatov et~al.(2020{\natexlab{a}})Ignatov, Timofte, Zhang, Liu, Wang, Zuo, Zhang, Zhang, Peng, Ren, et~al.]{ignatov2020aim2020challengelearned}
Andrey Ignatov, Radu Timofte, Zhilu Zhang, Ming Liu, Haolin Wang, Wangmeng Zuo, Jiawei Zhang, Ruimao Zhang, Zhanglin Peng, Sijie Ren, et~al.
\newblock Aim 2020 challenge on learned image signal processing pipeline.
\newblock In \emph{Computer Vision--ECCV 2020 Workshops: Glasgow, UK, August 23--28, 2020, Proceedings, Part III 16}, pages 152--170. Springer, 2020{\natexlab{a}}.

\bibitem[Ignatov et~al.(2020{\natexlab{b}})Ignatov, Van~Gool, and Timofte]{ignatov2020replacingmobilecameraisp}
Andrey Ignatov, Luc Van~Gool, and Radu Timofte.
\newblock Replacing mobile camera isp with a single deep learning model.
\newblock In \emph{Proceedings of the IEEE/CVF conference on computer vision and pattern recognition workshops}, pages 536--537, 2020{\natexlab{b}}.

\bibitem[Ignatov et~al.(2021)Ignatov, Chiang, Kuo, Sycheva, Timofte, et~al.]{ignatov2021learnedsmartphoneispmobile}
Andrey Ignatov, Cheng-Ming Chiang, Hsien-Kai Kuo, Anastasia Sycheva, Radu Timofte, et~al.
\newblock Learned smartphone isp on mobile npus with deep learning, {Mobile AI} 2021 challenge: Report.
\newblock In \emph{Proceedings of the IEEE/CVF Conference on Computer Vision and Pattern Recognition}, pages 2503--2514, 2021.

\bibitem[Ignatov et~al.(2022{\natexlab{a}})Ignatov, Malivenko, Timofte, Tseng, Xu, Yu, Chiang, Kuo, Chen, Cheng, et~al.]{ignatov2022pynetv2mobileefficientondevice}
Andrey Ignatov, Grigory Malivenko, Radu Timofte, Yu Tseng, Yu-Syuan Xu, Po-Hsiang Yu, Cheng-Ming Chiang, Hsien-Kai Kuo, Min-Hung Chen, Chia-Ming Cheng, et~al.
\newblock Pynet-v2 mobile: Efficient on-device photo processing with neural networks.
\newblock In \emph{2022 26th International Conference on Pattern Recognition (ICPR)}, pages 677--684. IEEE, 2022{\natexlab{a}}.

\bibitem[Ignatov et~al.(2022{\natexlab{b}})Ignatov, Sycheva, Timofte, Tseng, Xu, Yu, Chiang, Kuo, Chen, Cheng, et~al.]{ignatov2022microispprocessing32mpphotos}
Andrey Ignatov, Anastasia Sycheva, Radu Timofte, Yu Tseng, Yu-Syuan Xu, Po-Hsiang Yu, Cheng-Ming Chiang, Hsien-Kai Kuo, Min-Hung Chen, Chia-Ming Cheng, et~al.
\newblock Microisp: processing 32mp photos on mobile devices with deep learning.
\newblock In \emph{European Conference on Computer Vision}, pages 729--746. Springer, 2022{\natexlab{b}}.

\bibitem[Ignatov et~al.(2022{\natexlab{c}})Ignatov, Timofte, Liu, Feng, Bai, Wang, Lei, Yi, Xiang, Liu, et~al.]{ignatov2022learnedsmartphoneispmobile}
Andrey Ignatov, Radu Timofte, Shuai Liu, Chaoyu Feng, Furui Bai, Xiaotao Wang, Lei Lei, Ziyao Yi, Yan Xiang, Zibin Liu, et~al.
\newblock Learned smartphone isp on mobile gpus with deep learning, mobile ai \& aim 2022 challenge: report.
\newblock In \emph{European Conference on Computer Vision}, pages 44--70. Springer, 2022{\natexlab{c}}.

\bibitem[Ignatov et~al.(2025)Ignatov, Perevozchikov, Timofte, et~al.]{MAI2025learnedISP}
Andrey Ignatov, Georgii Perevozchikov, Radu Timofte, et~al.
\newblock Learned smartphone isp on mobile gpus, mobile ai 2025 challenge: Report.
\newblock In \emph{Proceedings of the IEEE/CVF Conference on Computer Vision and Pattern Recognition (CVPR) Workshops}, 2025.

\bibitem[Johnson et~al.(2016)Johnson, Alahi, and Fei-Fei]{johnson2016perceptuallossesrealtimestyle}
Justin Johnson, Alexandre Alahi, and Li Fei-Fei.
\newblock Perceptual losses for real-time style transfer and super-resolution.
\newblock In \emph{Computer Vision--ECCV 2016: 14th European Conference, Amsterdam, The Netherlands, October 11-14, 2016, Proceedings, Part II 14}, pages 694--711. Springer, 2016.

\bibitem[Jolicoeur-Martineau(2018)]{jolicoeurmartineau2018relativisticdiscriminatorkeyelement}
Alexia Jolicoeur-Martineau.
\newblock The relativistic discriminator: a key element missing from standard gan.
\newblock \emph{arXiv preprint arXiv:1807.00734}, 2018.

\bibitem[Kim et~al.(2020{\natexlab{a}})Kim, Song, Ye, and Baek]{Kim_2020}
Byung-Hoon Kim, Joonyoung Song, Jong~Chul Ye, and JaeHyun Baek.
\newblock Pynet-ca: enhanced pynet with channel attention for end-to-end mobile image signal processing.
\newblock In \emph{European Conference on Computer Vision}, pages 202--212. Springer, 2020{\natexlab{a}}.

\bibitem[Kim et~al.(2020{\natexlab{b}})Kim, Kim, Kang, and Lee]{kim2020ugatitunsupervisedgenerativeattentional}
Junho Kim, Minjae Kim, Hyeonwoo Kang, and Kwanghee Lee.
\newblock U-gat-it: Unsupervised generative attentional networks with adaptive layer-instance normalization for image-to-image translation.
\newblock In \emph{Proceedings of the 8th International Conference on Learning Representations (ICLR)}, 2020{\natexlab{b}}.

\bibitem[Krizhevsky et~al.(2017)Krizhevsky, Sutskever, and Hinton]{alexnet}
Alex Krizhevsky, Ilya Sutskever, and Geoffrey~E. Hinton.
\newblock Imagenet classification with deep convolutional neural networks.
\newblock \emph{Commun. ACM}, 60\penalty0 (6):\penalty0 84–90, 2017.

\bibitem[Land and McCann(1971)]{Land1971LightnessAR}
Edwin~Herbert Land and John~J. McCann.
\newblock Lightness and retinex theory.
\newblock \emph{Journal of the Optical Society of America}, 61 1:\penalty0 1--11, 1971.

\bibitem[Li et~al.(2024)Li, Hou, and Jia]{li2024rmfanetneuralispreal}
Fei Li, Wenbo Hou, and Peng Jia.
\newblock Rmfa-net: A neural isp for real raw to rgb image reconstruction.
\newblock \emph{arXiv preprint arXiv:2406.11469}, 2024.

\bibitem[Liu et~al.(2017)Liu, Breuel, and Kautz]{liu2018unsupervisedimagetoimagetranslationnetworks}
Ming-Yu Liu, Thomas Breuel, and Jan Kautz.
\newblock Unsupervised image-to-image translation networks.
\newblock \emph{Advances in neural information processing systems}, 30, 2017.

\bibitem[Low(2004)]{sift}
David~G Low.
\newblock Distinctive image features from scale-invariant keypoints.
\newblock \emph{Journal of Computer Vision}, 60\penalty0 (2):\penalty0 91--110, 2004.

\bibitem[Menon et~al.(2007)Menon, Andriani, and Calvagno]{4032820}
Daniele Menon, Stefano Andriani, and Giancarlo Calvagno.
\newblock Demosaicing with directional filtering and a posteriori decision.
\newblock \emph{IEEE Transactions on Image Processing}, 16\penalty0 (1):\penalty0 132--141, 2007.

\bibitem[Perevozchikov et~al.(2024)Perevozchikov, Mehta, Afifi, and Timofte]{perevozchikov2024rawformerunpairedrawtorawtranslation}
Georgy Perevozchikov, Nancy Mehta, Mahmoud Afifi, and Radu Timofte.
\newblock Rawformer: Unpaired raw-to-raw translation for learnable camera isps.
\newblock In \emph{European Conference on Computer Vision}, pages 231--248. Springer, 2024.

\bibitem[Raimundo et~al.(2022)Raimundo, Ignatov, and Timofte]{9856970}
Daniel~Wirzberger Raimundo, Andrey Ignatov, and Radu Timofte.
\newblock Lan: Lightweight attention-based network for raw-to-rgb smartphone image processing.
\newblock In \emph{2022 IEEE/CVF Conference on Computer Vision and Pattern Recognition Workshops (CVPRW)}, pages 807--815, 2022.

\bibitem[Simonyan and Zisserman(2015)]{simonyan2015deepconvolutionalnetworkslargescale}
Karen Simonyan and Andrew Zisserman.
\newblock Very deep convolutional networks for large-scale image recognition.
\newblock In \emph{Proceedings of the 3rd International Conference on Learning Representations (ICLR)}, 2015.

\bibitem[Truong et~al.(2021)Truong, Danelljan, Van~Gool, and Timofte]{truong2021learningaccuratedensecorrespondences}
Prune Truong, Martin Danelljan, Luc Van~Gool, and Radu Timofte.
\newblock Learning accurate dense correspondences and when to trust them.
\newblock In \emph{Proceedings of the IEEE/CVF conference on computer vision and pattern recognition}, pages 5714--5724, 2021.

\bibitem[Vedaldi and Fulkerson(2010)]{ransac}
Andrea Vedaldi and Brian Fulkerson.
\newblock Vlfeat: an open and portable library of computer vision algorithms.
\newblock In \emph{Proceedings of the 18th ACM International Conference on Multimedia}, page 1469–1472, New York, NY, USA, 2010. Association for Computing Machinery.

\bibitem[Wang et~al.(2003)Wang, Simoncelli, and Bovik]{wang2003multiscale}
Zhou Wang, Eero~P Simoncelli, and Alan~C Bovik.
\newblock Multiscale structural similarity for image quality assessment.
\newblock In \emph{The Thrity-Seventh Asilomar Conference on Signals, Systems \& Computers, 2003}, pages 1398--1402. Ieee, 2003.

\bibitem[Wang et~al.(2004)Wang, Bovik, Sheikh, and Simoncelli]{wang2004image}
Zhou Wang, Alan~C Bovik, Hamid~R Sheikh, and Eero~P Simoncelli.
\newblock Image quality assessment: from error visibility to structural similarity.
\newblock \emph{IEEE transactions on image processing}, 13\penalty0 (4):\penalty0 600--612, 2004.

\bibitem[Zhang et~al.(2018)Zhang, Isola, Efros, Shechtman, and Wang]{zhang2018unreasonableeffectivenessdeepfeatures}
Richard Zhang, Phillip Isola, Alexei~A Efros, Eli Shechtman, and Oliver Wang.
\newblock The unreasonable effectiveness of deep features as a perceptual metric.
\newblock In \emph{Proceedings of the IEEE conference on computer vision and pattern recognition}, pages 586--595, 2018.

\bibitem[Zhu et~al.(2017)Zhu, Park, Isola, and Efros]{zhu2020unpairedimagetoimagetranslationusing}
Jun-Yan Zhu, Taesung Park, Phillip Isola, and Alexei~A Efros.
\newblock Unpaired image-to-image translation using cycle-consistent adversarial networks.
\newblock In \emph{Proceedings of the IEEE international conference on computer vision}, pages 2223--2232, 2017.

\end{thebibliography}
}


\end{document}